# Markov chain Hebbian learning algorithm with ternary synaptic units

Guhyun Kim[1,2], Vladimir Kornijcuk[1,3], Dohun Kim[1,2], Inho Kim[1], Jaewook Kim[1], Hyo Cheon Woo,[2] Ji Hun Kim,[2] Cheol Seong Hwang[2]*, and Doo Seok Jeong[1,3]*

[1]*Center for Electronic Materials, Korea Institute of Science and Technology, Hwarangno 14-gil 5, Seongbuk-gu, 02792 Seoul, Republic of Korea*

[2]*Department of Materials Science and Engineering and Inter-University Semiconductor Research Centre, Seoul National University, 151-744 Seoul, Republic of Korea*

[3]*Department of Nanomaterials, Korea University of Science and Technology, Daejeon, Republic of Korea*

In spite of remarkable progress in machine learning techniques, the state-of-the-art machine learning algorithms often keep machines from real-time learning (online learning) due in part to computational complexity in parameter optimization. As an alternative, a learning algorithm to train a memory in real time is proposed, which is named as the Markov chain Hebbian learning algorithm. The algorithm pursues efficient memory use during training in that (i) the weight matrix has ternary elements (-1, 0, 1) and (ii) each update follows a Markov chain—the upcoming update does not need past weight memory. The algorithm was verified by two proof-of-concept tasks (handwritten digit recognition and multiplication table memorization) in which numbers were taken as symbols. Particularly, the latter bases multiplication arithmetic on memory, which may be analogous to humans' mental arithmetic. The memory-based multiplication arithmetic feasibly offers the basis of factorization, supporting novel insight into the arithmetic.

*Address correspondence to: C.S.H. (cheolsh@snu.ac.kr) or D.S.J. (dsjeong@kist.re.kr)

Recent progress in machine learning (particularly, deep learning) endows artificial intelligence with high precision recognition and problem-solving capabilities beyond the human level.[1, 2, 3] Computers on the von Neumann architecture are the platform for the breakthroughs albeit frequently powered by hardware accelerators, e.g. graphics processing unit (GPU).[4] The main memory, in this case, is used to store fragmentary information, e.g. weight matrix, representation of hidden neurons, and input datasets, intertwined among the fragments. Therefore, it is conceivable that memory organization is essential to efficient memory retrieval. To this end, memory keeping a weight matrix in place can be considered, in which the matrix matches different representations selectively as a consequence of learning. For instance, a visual input (representation) such as a hand-written '1' recalls a symbolic memory '1' (internal representation) through the stored weight matrix so that the symbolic memory can readily be recalled. In this regard, a high-density crossbar array (CBA) of two-terminal memory elements, e.g. oxide-based resistive memory and phase-change memory, is perhaps a promising solution to machine learning acceleration.[5, 6, 7, 8, 9] The connection weight between a pair of neurons is stored in each memory element in the CBA as conductance, and the weight is read out in place by monitoring current in response to a voltage.[5, 6, 7, 8, 9] Albeit promising, this approach should address the following challenges; each weight should be pre-calculated beforehand using a conventional error-correcting technique, and the pre-calculated value state needs to be accommodated by a single memory element. The former particularly hinders online learning.

In this study, an easy-to-implement algorithm based on a stochastic neural network—termed as the Markov chain Hebbian learning (MCHL) algorithm is proposed. The most notable difference between the MCHL and restricted Boltzmann machine (RBM)[10, 11, 12, 13] is that the MCHL is a discriminative learning algorithm with the aid of "external field" realizing supervised learning. Also, each update concerns only individual neuron-to-neuron weight values rather than the energy of the entire network, i.e. greedy edge-wise training. The MCHL algorithm also features as follows:

(a) Each weight $w[i, j]$ is a ternary number: $w[i, j] \in \{-1, 0, 1\}$.
(b) Given (a), each update of weight follows a finite-state Markov chain, and the update probability is in line with the Hebbian learning.

(c) A group of output neurons in a bucket (rather than a single neuron) simultaneously represent a data class (label), which is comparable to concept cells[14, 15, 16].

(d) When the network is deep, the network is trained in a greedy layer-wise manner, and each layer is trained in a greedy edge-wise manner.

Provided with these features, the MCHL algorithm enables an *ad hoc* update of the weight matrix (online learning) in a memory-saving fashion, so that it is suitable for machine learning powered by CBA-based memory. No need for an auxiliary function for error correction, e.g. backpropagation, particularly alleviates computational complexity, and thus severe needs for massive computing power are eliminated. Additionally, each synapse is given a ternary number during the entire learning period—distinguishable from binarizing real-valued weight at each update step[17] as well as the use of auxiliary real-valued variables[18].

The MCHL algorithm was applied to two proof-of-concept examples: handwritten digit recognition using the MNIST database and multiplication table memorization in the framework of supervised learning for classification. The latter example relates the arithmetic to memory-based perception in an analogous way to humans' mental arithmetic. The weight matrix trained with the multiplication table was then applied to more complicated arithmetic such as aliquot part evaluation and prime factorization.

**Results**

**Basic network structure and its energy.** Analogous to the RBM, two layers of binary stochastic neurons without recurrent connection form the basis for the MCHL algorithm. However, it differs from the RBM such that the hidden layer in the RBM is replaced by an output layer that does not feed input into the input layer. Figure 1**a** depicts a stochastic neural network of $M$ input neurons and $N$ output neurons; $u_1$ and $u_2$ denote the activity vector of the first and second layer, respectively. Note that $u_1 \in \mathbb{R}^M$ ($0 \leq u_1[i] \leq 1$), unless otherwise specified. $u_2 \in \mathbb{Z}^N$; $u_2[i] \in \{0, 1\}$. The energy of this network is defined as

$$E(\boldsymbol{u}_1, \boldsymbol{u}_2) = -(2\boldsymbol{u}_2 - \vec{\boldsymbol{1}})^T \cdot \boldsymbol{w} \cdot \boldsymbol{u}_1 + \boldsymbol{a}^T \cdot \boldsymbol{u}_2, \qquad \text{Eq. (1)}$$

where $\boldsymbol{w}$ is a weight matrix ($\boldsymbol{w} \in \mathbb{Z}^{N \times M}$; $w[i,j] \in \{-1, 0, 1\}$), and $\vec{\boldsymbol{1}}$ is a $N$-long vector filled with ones. $\boldsymbol{a}$ denotes a bias vector for the output layer. $\boldsymbol{a}^T$ is the transpose of a vector $\boldsymbol{a}$. Given that $2u_2[i] - 1 \in \{-1, 1\}$, $(2\boldsymbol{u}_2 - \vec{\boldsymbol{1}})$ in Eq. (1) effectively transforms $\boldsymbol{u}_2$ in a way that a quiet neuron ($u_2[i] = 0$) is given an output of -1 rather than zero. This counts the cost of a positive connection ($w[i, j] = 1$) between a nonzero input ($u_1[j] \neq 0$) and output neuron in an undesired label ($u_2[i] = 0$) in supervised learning; this undesired connection raises the energy by $u_1[j]$. The summation of inputs into the $i$th output neuron is given by $z[i] = \sum_{j=1}^{M} w[i,j] u_1[j]$. The following conditional probability that $u_2[i] = 1$ given $z[i]$ holds:

$$P(u_2[i] = 1 | z[i]) = \frac{1}{1 + e^{-(2z[i] - a[i])/\tau}}. \qquad \text{Eq. (2)}$$

$\tau$ denotes a temperature parameter. Eq. (2) is plotted in Fig. 1**b**. The derivation of Eq. (2) is elaborated in **Methods**. Note that unless otherwise stated, the bias is set to zero, simplifying Eqs. (1) and (2) to $E(\boldsymbol{u}_1, \boldsymbol{u}_2) = -(2\boldsymbol{u}_2 - \boldsymbol{1})^T \cdot \boldsymbol{w} \cdot \boldsymbol{u}_1$ and $P(u_2[i] = 1 | z[i]) = 1/(1 + e^{-2z[i]/\tau})$, respectively.

**Field application and update probability.** In the MCHL algorithm, supervision is realized by applying a field that directs input pattern to its desired label. Directing input is implemented by (a) encouraging its connection with an output neuron(s) with the desired label among $L$ labels and (b) discouraging otherwise—both in a probabilistic manner. To this end, a write vector $\boldsymbol{v}$ that points to the correct label in the $L$-dimensional space is essential. Each label is given a bucket of $H$ neurons so that $\boldsymbol{v}$ is an $LH$-long vector. Given that all labels are orthogonal to each other, each bucket of $\boldsymbol{v}$, i.e. $v[(n-1)H+1:nH]$; $1 \leq n \leq L$, offers each basis of the applied field in the $L$-dimensional space. $v[a:b]$ denotes a block ranging from the $a$th to $b$th elements. Only one element in each label bucket of $\boldsymbol{v}$ is randomly chosen for each *ad hoc* update and given non-zero value in that the element dedicated to the desired label is set to 1 while the other $L$-1 elements to -1. This write vector $\boldsymbol{v}$ is renewed every update. Therefore, the update is *sparse*. It is noteworthy that $v[i] \in \{-1, 1\}$ when $H = 1$ and $v[i] \in \{-1, 0, 1\}$ otherwise.

Figure 1**c** graphically describes the feed-forward connection between $u_1$ and $u_2$ for the topology in Fig. 1**a**. The matrix $w$ is loaded with ternary elements ($w \in \mathbb{Z}^{N \times M}$; $w[i,j] \in \{-1, 0, 1\}$ and $N=LH$). Here the input vector $u_1 \in \mathbb{R}^M$; $0 \leq u_1[i] \leq 1$. Consequently, $v \in \mathbb{Z}^N$; $v[i] \in \{-1, 0, 1\}$. According to the bucket configuration of the write vector $v$, the matrix $w$ is partitioned such that $w[(n-1)H+1:nH, 1:M]$ defines the correlation between the input and its label ($n$). Likewise, $z$ ($= wu_1$) is also partitioned into $H$ buckets in the same order as $v$, and the same holds for the output activity vector $u_2$.

Every pair of $u_1$ and $v$ stochastically updates each component $w[i, j]$ in $w$ by $\Delta w[i,j] = w_{t+1}[i,j] - w_t[i,j] \in \{-1, 0, 1\}$. The variables determining $\Delta w[i, j]$ include (a) $u_1[j]$ and $v[i]$, (b) current value of $w_t[i, j]$, and (c) output activity $u_2[i]$ as follows (also see Table 1).

**Condition (a)**: it is probable that $\Delta w[i, j] = 1$ when $u_1[j]v[i] > 0$ (i.e. $u_1[j] \neq 0$ and $v[i] = 1$) and $\Delta w[i, j] = -1$ when $u_1[j]v[i] < 0$ (i.e. $u_1[j] \neq 0$ and $v[i] = -1$) conditional upon (b) and (c). That is, $w[i, j]$ is updated to connect the nonzero $u_1[j]$ and $i$th output neuron in the desired label (when $v[i] = 1$) and to disconnect when $v[i] = -1$. The former and latter updates are referred to as potentiation and depression, respectively (Figs. 1**d** and **e**). This condition is reminiscent of the Hebbian learning such that $\Delta w[i, j]$ is determined by $u_1[j]v[i]$. The larger the input $u_1[j]$, the more likely the update is successful such that both $P_+$ (potentiation probability) and $P_-$ (depression probability) scale with $u_1[j]$; $P_+ = u_1[j]P_+^0$ and $P_- = u_1[j]P_-^0$, where $P_+^0$ and $P_-^0$ denote the maximum probability of potentiation and depression, respectively. Such a negative update is equivalent to homosynaptic long-term depression in the biological neural network, elucidated by the Bienenstock-Cooper-Munro theory supporting the spontaneous selectivity development.[19, 20]

**Condition (b)**: The updates $\Delta w[i, j] = 1$ and $\Delta w[i, j] = -1$ given Condition (a) are allowed if the current weight is not 1 ($w_t[i, j] \neq 1$) and not -1 ($w_t[i, j] \neq -1$), respectively. This condition keeps $w[i,j] \in \{-1, 0, 1\}$ so that the update falls into a finite-state Markov chain.

**Condition (c)**: Alongside Conditions (a) and (b), the updates $\Delta w[i, j] = 1$ and $\Delta w[i, j] = -1$ require $u_2[i] = 0$ and $u_2[i] = 1$, respectively. That is, a quiet output neuron ($u_2[i] = 0$) supports $\Delta w[i, j] = 1$, whereas an active one ($u_2[i] = 1$) supports $\Delta w[i, j] = -1$.

As a consequence of these update conditions, the MCHL algorithm spontaneously captures the correlation between input and write vectors ($u_1$ and $v$) during repeated Markov processes, which is exemplified in Supplementary Information for randomly generated input and write vectors that have a statistical correlation.

As such, a learning rate is of significant concern for successful learning; a proper rate that allows the matrix to converge to the optimized one should be chosen. The same holds for the MCHL algorithm. The rate in the proposed algorithm is dictated by $P_+^0$ and $P_-^0$ in place of an explicit rate term. For extreme cases such as $P_+^0 = 1$ and $P_-^0 = 1$, the matrix barely converges, but constantly fluctuates.

**Handwritten digit recognition (supervised online learning).** The MCHL algorithm was applied to the handwritten digit recognition task with the MNIST database ($L = 10$). Figure 2**a** shows the network schematic for the training, which includes one hidden layer. This network is trained in a greedy layer-wise manner as for deep belief networks[21]. That is, the matrix $w_1$ was first fully trained with an input vector $u_1$ and write vector $v_1$, which was then followed by training the matrix $w_2$ with $P(u_2 = 1)$ and $v_2$. $P(u_2 = 1)$ is a vector of probability that $u_2[i] = 1$ (see Eq. (2)). $P(u_2 = 1)$ in response to each MNIST training example was taken as the input to the matrix $w_2$. The training protocol was the same for both matrices

The network depth substantially alters the recognition accuracy as plotted in Fig. 2**b**. Without hidden layer (HL) the accuracy merely reaches approximately 88% at $H_1 = 100$ while deploying one HL improves the accuracy up to approximately 92% at $H_1 = 100$ and $H_2 = 50$. Improvement on accuracy continues onwards with more HLs (e.g. two HLs; blue curve in Fig. 2**b**), although its effect becomes smaller compared with the drastic improvement by the first HL. The training and test in detail are addressed in **Methods**.

The weight matrix becomes larger with bucket size, so is the memory allocated for the matrix. Nevertheless, the benefit of deploying buckets at the expense of memory is two-fold. First, many input features (pixels) are shared among labels such that an individual feature should not exclusively belong to a single particular label. The use of buckets allows such common features to be connected with elements over different labels given the *sparse* update on the weight matrix. For instance, without such buckets, every attempt to direct the feature at (1, 1) —belonging to both labels 1 and 2— to label 1 probabilistically weakens its connection with label 2. Second, when shared, the statistical correlation between the feature and each of the sharing labels is captured by bucket, enabling comparison among the labels. As depicted in Fig. 2**a**, the 10 sub-matrices in the matrix $w_2$ define 10 ensembles of $H_2$ output neurons; the final output from each label $O[i]$ is the sum of output over the neurons in the same label, i.e. the output range scales with $H_2$ in the range $0 - H_2$. A single training is hardly able to capture the statistical correlation between the input and write vectors due to a large error. However, the larger the trial numbers, the less likely the statistical error (noise) is incorporated into the data, which is similar to the error reduction in Monte Carlo simulation with an enormous number of random numbers (RNs).[22] The use of buckets enables the parallel acquisition of effectively multiple $w$ matrices as opposed to repeated training trials to acquire a $w$ matrix on average as for the toy example in Supplementary Fig. 1. Therefore, it is conceivable that a larger bucket size tends to improve the recognition accuracy. In fact, the bucket size and consequent memory allocation for matrix $w$ significantly determine the recognition accuracy (see Supplementary Fig. 2). However, in Monte Carlo simulations, the error reduction with sample number tends to be negligible when the number is sufficiently large. The same holds for the MCHL algorithm as shown in Supplementary Fig. 2. Additionally, the memory cost perhaps outweighs the negligible improvement in the accuracy. Therefore, it is practically important to reconcile the performance with the memory cost.

**Multiplication table memorization.** The MCHL algorithm can also be applied to deterministic learning. Examples include multiplication table memorization in which the MCHL algorithm spontaneously finds the correct-answer-addressing matrix $w$. This way allows addressing correct answers, rather than performing multiplication. A matrix $w$ ($w \in \mathbb{Z}^{N \times 2M}$; $w[i,j] \in \{0,1\}, N = M^2 H$) was subject to training with the $M \times M$ multiplication table. Two integer factors in the range $(1 - M)$ were chosen and represented by two one-hot vectors, each of which had $M$ elements. These two vectors were merged into an input vector $u_1$ ($\in \mathbb{Z}^{2M}$; $u_1[i] \in \{0,1\}$); $u_1[1:M]$ were allocated for the first vector, and $u_1[M+1:2M]$ for the second one. The product ranges from 1 to $M^2$—which is taken as the desired label of the input. Therefore, $M^2$ labels in total are available. Given bucket size $H$ for each label, the write vector $\boldsymbol{v}$ is $M^2H$ long.

Multiplication is deterministic so that no stochasticity intervenes in learning. Consequently, $P_+^0 = 1$ and $P_-^0 = 0$ were given to Eq. (7), and all neurons were frozen ($\tau = 0.01$). In this regard, write vector generation does not require random sampling within the bucket in the desired label. Instead, an element in the bucket is conferred on each pair of factors in order of training. For instance, $2 \times 8$ addresses the $n$th element in label 16, and the multiplication addressing the same label in the closest succession, e.g. $4 \times 4$, takes the ($n$+1)th element. Therefore, the bucket includes a set of possible multiplications that yields the same label. Notably, a prime number has only two factors, '1' and itself, and thus, the bucket includes only two multiplications. Note that bias is given to each output neuron; $a[i] = 3$ for all $i$'s. Therefore, Eq. (2) in this application is expressed as $P(u_2[i] = 1|z[i]) = 1/\left[1 + e^{-(2z[i]-3)/\tau}\right]$. The bias allows $u_2[i] = 1$ only if $z[i] > 2$ so that a single factor cannot solely activate the output neuron.

The network structure is sketched in Fig. 3**a**; no HL is required to achieve the maximum accuracy. The training continued onwards until the entire pairs of numbers in the table were subject to memorization. $M^2$ training steps were thus required to complete the memorization task. An indexing vector $A$ $\left(\in \mathbb{Z}^{M^2}; A[i] = h\right)$ was defined to count the possible multiplications ($h$) resulting in the same product. For instance, when $M \geq 6$, $A[6] = 4$ because $1 \times 6$, $2 \times 3$, $3 \times 2$, and $6 \times 1$ result in 6 (see Fig. 3**a**). Notably, A[$i$] is identical to the number of factors of $i$. The training procedure is elaborated in **Methods**. Note that the prime numbers large than $M$ cannot be taken as a label.

The trained matrix $\boldsymbol{w}$ can readily be used to find the aliquot parts of number $n$ by transposing the matrix: $\boldsymbol{w}^T \in \mathbb{Z}^{2M \times N}; N = M^2H$ (see Fig. 3**b**). The matrix multiplication $\boldsymbol{z} = \boldsymbol{w}^T\boldsymbol{u}_1$ with $\boldsymbol{u}_1$ ($\in \mathbb{Z}^N; N = M^2H$)—all $H$ elements in the $n$th bucket are set to 1—yields a vector $\boldsymbol{z}$ whose upper $M$ bits $z[1:M]$ are the sum of the entire aliquot parts, each of which is represented by a one-hot vector (Fig. 3**b**). Given the commutative property of multiplication, $z[1:M] = z[M +1:2M]$. For instance, when $M$

= 9, input '6' yields $z[1:9]$ = [111001000], indicating '1' + '2' + '3' + '6'. A prime number '7' yields $z[1:9]$ = [100000100] ('1' + '7'); two 1's in $z$ indicates a prime number ($h = 2$).

**Prime factorization.** The matrix $w$ trained with an $M \times M$ multiplication table also serves as the basis for prime factorization (Fig. 4**a**). It is a modified version of the aliquot part retrieval to avoid retrieving '1' and itself if other factors exist. A remarkable advantage consists in the parallel decomposition of many numbers; for an input $u$ (the sum of one-hot vectors under decomposition, e.g. A = a × b and B = c × d), the single matrix-vector multiplication $z = w^T u$ uncovers all a, b, c, and d.

Figure 4**b** illustrates a factor tree of '840'; the first iteration with $w$ ($M = 50$) results in '40' + '21', the following iteration gives '2' + '3' + '7' + '20', and the third iteration 2×'2' + '3' + '7' + '10', equivalent to $a_1$, $a_2$, and $a_3$ in Fig. 4**c**. To demonstrate the efficiency of this method, randomly picked integers in a multiplication table ($M$=300) were subject to prime factorization, and the number of the iteration steps was counted. The results for 25 integers are plotted in Fig. 4**d** in comparison with a benchmark (direct search factorization). The higher efficiency of the present method over the benchmark can obviously be understood. The direct search factorization is elaborated in **Methods**.

The capacity for prime factorization using the proposed algorithm is dictated by the size of a trained $M \times M$ multiplication table. As such, the larger the size $M$, the more the factorizable integers (Supplementary Fig. 3). Note that the factorizable integers should be addressed as a product in the $M \times M$ multiplication table such that the number of factorizable integers is identical to that of products in the table. There exist 36 different products in the 9×9 multiplication table; all of them are prime-factorizable. Upon enlarging the table size up to $M$=300, the capacity reaches 24,047. Given the ternary weight in $w$ (each element needs 2 bits), the required memory size for $w$ ($M$=300) is 180 kbits (Supplementary Fig. 3).

## Discussion

The MCHL algorithm employs the population representation of output neurons; the population is partitioned as a consequence of bucket allocation for each label. This notion is reminiscent of 'concept cells'.[14, 15, 16] They fire only to specific inputs that point to the same concept even with different stimulus modalities.[15] Likewise, the 10 populations in Fig. 2**a** may be equivalent to concept cells, each of which represents each digit. Additionally, deploying buckets may support the integration of different stimulus modalities, each of which is directed to the same concept cell throughout different pathways. This bucket can include different neurons at the pinnacles of different pathways, e.g. in an auditory modality, so that these different stimulus modalities can complementarily activate the bucket.

Given that each bucket represents a single concept, a one-hot vector representation is most suitable for the mathematical description of concepts. The proposed multiplication table memorization algorithm therefore lays the foundation of arithmetic in association with perception via memory. All integers (factors and products) in the table are represented by one-hot vectors that are equivalent to concept cells. They may be addressed by not only arithmetic but also external stimuli in different sensory modalities. Arithmetic with the aid of memory may be akin to humans' mental arithmetic, particularly, of simple single-digit arithmetic.[23, 24, 25] Additionally, this memory-based multiplication may combine arithmetic with different sensory modalities, e.g. multiplication triggered by voice commands and/or handwritten problem sets.

The MCHL algorithm offers a solution to online learning given that the algorithm enables *ad hoc* updates on a weight matrix accommodated by a random access memory (RAM) without pre-calculating the weight matrix. This approach, therefore, provides a workaround for the matrix calculation overhead that is a challenge when addressing representations with enormous features. Additionally, the ternary (-1, 0, 1) weight elements—each of which merely needs two bits—significantly improve the areal density of the matrix mapped onto a RAM array in support of density- as well as the energy-wise efficiency of training. A CBA of resistance-based memory is perhaps most suitable for the MCHL algorithm, which emphasizes the benefits of the MCHL algorithm for efficient matrix calculation.[5, 9, 26] Given the stochasticity in resistance switching (particularly, on- and off-switching voltages[27, 28]) in nature, the probabilistic weight transition may be achieved by controlling driving voltage without RN generation[29]. Additionally, every update simply overwrites the current memory contents in this training scheme in that the past weight matrix no longer needs to be kept given the Markov chain nature, which also alleviates large memory needs.

**Methods**

**Derivation of stochastic activity of a neuron.** Given the network energy in Eq. (1), the joint probability distribution of the state $u_1$ and $u_2$ is described as $P(\boldsymbol{u}_1, \boldsymbol{u}_2) = e^{-E(\boldsymbol{u}_1, \boldsymbol{u}_2)/\tau}/Z$, where $Z$ is the partition function of the network, $Z = \sum_{j=1}^{M} \sum_{i=1}^{N} e^{-E(u_1[j], u_2[i])/\tau}$. Consequently, the conditional probability distribution of $u_2$ given $u_1$ is

$$P(\boldsymbol{u}_2|\boldsymbol{u}_1) = \frac{e^{\sum_{i=1}^{N}\left(-a[i]u_2[i] - \sum_{j=1}^{M} w[i,j]u_1[j] + 2\sum_{j=1}^{M} u_2[i]w[i,j]u_1[j]\right)/\tau}}{\prod_{i=1}^{N} \sum_{u_2[i] \in \{0,1\}} e^{\left(-a[i]u_2[i] - \sum_{j=1}^{M} w[i,j]u_1[j] + 2\sum_{j=1}^{M} u_2[i]w[i,j]u_1[j]\right)/\tau}} =$$

$$\prod_{i=1}^{N} \frac{e^{\left(-a[i]u_2[i] + 2\sum_{j=1}^{M} u_2[i]w[i,j]u_1[j]\right)/\tau}}{1 + e^{\left(-a[i] + 2\sum_{j=1}^{M} w[i,j]u_1[j]\right)/\tau}}. \qquad \text{Eq. (3)}$$

$P(\boldsymbol{u}_2|\boldsymbol{u}_1) = \prod_{i=1}^{N} P(u_2[i]|\boldsymbol{u}_1)$ such that $u_2[i]$'s are independent of each other owing to the lack of recurrent connection. Therefore, the following equation holds:

$$P(u_2[i] = 1|\boldsymbol{u}_1) = \frac{e^{\left(-a[i] + 2\sum_{j=1}^{M} w[i,j]u_1[j]\right)/\tau}}{1 + e^{\left(-a[i] + 2\sum_{j=1}^{M} w[i,j]u_1[j]\right)/\tau}}. \qquad \text{Eq. (4)}$$

Introducing $z[i]\left(= \sum_{j=1}^{M} w[i,j]u_1[j]\right)$ simplifies Eq. (4) to

$$P(u_2[i] = 1|z[i]) = \frac{e^{(-a[i] + 2z[i])/\tau}}{1 + e^{(-a[i] + 2z[i])/\tau}} = \frac{1}{1 + e^{(a[i] - 2z[i])/\tau}}. \qquad \text{Eq. (5)}$$

**Calculation of update probability.** The update conditions and corresponding probability $P$ can readily be incorporated into the following equation (when $v[i] \neq 0$):

$$P(\Delta w[i,j] = v[i]|w_t[i,j], u_1[j], v[i], u_2[i]) = \frac{u_1[j]v[i]\left[P_+^0(1 - u_2[i])(v[i] + 1) + P_-^0 u_2[i](v[i] - 1)\right]}{2\left[1 + e^{k(w_t[i,j]v[i] - w_0)}\right]}, \quad \text{Eq. (6)}$$

where $P_+^0$ and $P_-^0$ are expressed as $P(\Delta w[i,j] = 1 | w_t[i,j] \neq 1, u_1[j] = 1, v[i] = 1, u_2[i] = 0)$ and $P(\Delta w[i,j] = -1 | w_t[i,j] \neq -1, u_1[j] = 1, v[i] = -1, u_2[i] = 1)$, respectively. $k$ and $w_0$ dictate the exponential function in the denominator, which are set to 100 and 0.5 through the entire simulation.

The probability given various circumstances is tabulated in Supplementary Table 1. Note that when $v[i] = 0$ no update on $w[i,j]$ is allowed, i.e. $P(\Delta w[i,j] = 0 | v[i] = 0) = 1$.

In practical computation, the stochastic variable $u_2[i]$ with the probability in Eq. (2) is acquired with the aid of a single RN before applying Eq. (6) to the $w[i, j]$ update that needs another RN. Fortunately, $u_2[i]$ can be ruled out among the conditions in Eq. (6) as follows:

$$P(\Delta w[i,j] = v[i]|w_t[i,j], u_1[j], v[i]) = P(\Delta w[i,j] = v[i]|w_t[i,j], u_1[j], v[i], u_2[i] = 1)P(u_2[i] = 1) + P(\Delta w[i,j] = v[i]|w_t[i,j], u_1[j], v[i], u_2[i] = 0)P(u_2[i] = 0) =$$

$$\frac{u_1[j]v[i]\left[P_+^0(v[i]+1)e^{-(2z[i]-a[i])/\tau} + P_-^0(v[i]-1)\right]}{2[1+e^{k(w_t[i,j]v[i]-w_0)}]\left(1+e^{-(2z[i]-a[i])/\tau}\right)}. \quad \text{Eq. (7)}$$

Each update of $w[i, j]$, therefore, needs a single RN, rendering the computation more efficient.

**Handwritten digit recognition.** For the entire datasets, each feature value was rescaled to the range 0 – 1. A chosen input dataset (28 × 28 pixels) was converted to an input vector $\boldsymbol{u}_1$ ($\in \mathbb{R}^{784}; 0 \leq u_1[i] \leq 1$). A write vector $\boldsymbol{v}$ ($\in \mathbb{Z}^{LH}; v[i] \in \{-1, 0, 1\}$) was then generated with regard to the desired label of the chosen digit and RN $r$ ($1 \leq r \leq H$). $L$ and $H$ are the number of total labels (here 10) and bucket size, respectively. A bucket of $H$ elements is assigned to each label in the $\boldsymbol{v}$ vector so that $\boldsymbol{v}$ is a 10$H$-long vector as illustrated in Fig. 2**a**. Accordingly, the matrix $\boldsymbol{w}$ is partitioned into 10 sub-matrices. One of the $H$ elements ($r$th element) in the bucket of the correct label is chosen at random and set to 1, the $r$th elements in the other buckets (9 in total) to -1, and the rest elements [10($H$ - 1) in total] to 0. Therefore, in the matrix $\boldsymbol{w}$, the elements in only one row ($r$th row in the partition for the correct label) are potentially subject to potentiation, those in the 9 rows to depression ($r$th rows in the partitions for the incorrect labels), and the rest are invariant. The update is therefore sparse.

The weight matrices $\boldsymbol{w}$ was initially filled with zeros. The update direction and probability were determined by Eq. (7). Each *ad hoc* update needs total 784$LH$ RNs (one for each $w[i,j]$). The protocol was repeated for the next epoch with a randomly chosen digit. For accuracy evaluation, a vector $\boldsymbol{z}$ (= $\boldsymbol{wu}_1$) was calculated after every *ad hoc* update and fed into the output neurons that are also partitioned according to the bucket configuration in the write vector and weight matrix. Note that this accuracy evaluation no longer needs stochastic neurons since their probabilistic behaviour rather limits the accuracy. Thus, they are switched to sigmoid deterministic neurons only for accuracy evaluation, which follows $u_2[i] = 1/[1 + e^{-2z[i]/\tau}]$. Finally, the output from each label $i$ ($O[i]$) is given by the sum of neuronal output over the all neurons in the bucket of label $i$. The maximum component of the output vector designates the estimated label for a given input. The recognition accuracy was evaluated with regard to agreement between the desired and estimated labels. The sequence of the MCHL algorithm application is elaborated in Supplementary Table 2.

A network with a hidden layer is trained in a greedy layer-wise manner as for deep belief networks[21]. $w_1$ in Fig. 2a was first fully trained following the protocol above. Subsequently, $w_2$ was subject to training with input vector $u_2$ ($\in \mathbb{Z}^{LH_1}$; $u_2[i] \in \{0, 1\}$) that is the output from the $LH_1$ hidden deterministic neurons taking $z_1$ as input. The write vector $v_2$ was chosen applying the same protocol as $w_1$ training. Accuracy evaluation was conducted with deterministic sigmoid output neurons in line with the network without hidden layer.

**Multiplication table memorization**. Training was fully deterministic in that the output neurons were frozen and the update no longer required RNs. Integers ($\leq M$) were expressed as one-hot vectors of $M$ elements; a pair of factors ($\leq M$) were put together to give an input vector $u_1$ ($\in \mathbb{Z}^{2M}$; $u_1[i] \in \{0, 1\}$). The product of the factors serves as a label among $M^2$ labels, each of which has a bucket of $H$ elements. Therefore, a write vector $v$ has $M^2H$ elements in total ($v \in \mathbb{Z}^{M^2H}$; $v[i] \in \{0, 1\}$). For factors of $a$ and $b$ ($a \times b = c$), the $h$th element in the $c$th label, i.e. $v[(c-1), H+h]$, is set to the only one in the write vector. $h$ is determined in the order of training; the first pair of factors resulting in a particular label during training takes $h = 1$ in the corresponding bucket. Thus, allocating $h$ for each multiplication depends on the entire training sequence over the $M \times M$ multiplication table. The weight matrix $w$ ($\in \mathbb{Z}^{M^2H \times 2M}$; $w[i,j] \in \{0, 1\}$) was trained in an ascending order of $n$ in the $n$-times table ($n \times$) from 1 to $M$, and within the $n$-times table ($n \times m$), $m$ was also taken in ascending order: $1 \times 1$, $1 \times 2$, …, $1 \times M$, $2 \times 1$, $2 \times 2$, …, $2 \times M$, … $M \times 1$, $M \times 2$, … $M \times M$. Upon training completion, final $h$ ($\leq H$) for label $i$ (i.e. $h_i$) is acquired, which defines vector $A$ ($\in \mathbb{Z}^{M^2}$; $A[i] = h_i$). In fact, $A[i]$ reveals the number of multiplications producing label $i$, for instance, $A[6] = 4$ given that $1 \times 6$, $2 \times 3$, $3 \times 2$, and $6 \times 1$ result in 6 (see Fig. 3a). Notably, this number is identical to the number of factors for a given label: 1, 2, 3, and 6 for 6. The sequence of the MCHL algorithm application is tabulated in Supplementary Table 3.

$z[i]$ in $z$ (= $wu_1$) was integrated over elements in the bucket of each label, which was subsequently fed into an output sigmoid neuron, resulting in output vector $O$ as illustrated in Fig. 3a.

**Prime factorization.** As such, the aliquot parts of number $n$ are in parallel retrieved using the transpose of $w$ [$w^T \in \mathbb{Z}^{2M \times M^2 H}$] memorizing the $M \times M$ multiplication table and input vector $u$ ($\in \mathbb{Z}^{M^2 H}$; $u[i] \in \{0, 1\}$) whose $n$th bucket is filled with $H$ 1's—insofar as $n$'s largest aliquot part is not larger than $M$. However, for prime factorization of $n$, aliquot parts other than 1 and itself (if they exist) are of concern, so that it is desirable to avoid retrieving $1 \times n$ and $n \times 1$. With the aid of vector $A$, a pair of proper factors can be chosen selectively. As shown in Fig. 3a, for 6 ($M \geq 6$), $h$ = 1, 2, 3, and 4 indicate $1 \times 6$, $2 \times 3$, $3 \times 2$, and $6 \times 1$, respectively. For a prime number, e.g. 7, $h$ = 1 and 2 indicate $1 \times 7$ and $7 \times 1$, respectively. Only the $k$th multiplication is retrieved, $k = \max(A[i] - 1, 1)$ for each label $i$, e.g. for $i$ = 6 ($M \geq 6$), $3 \times 2$, and for $i$ = prime number ($M \geq n$), $1 \times n$. Thus, operator $T_1$ is a $M^2 H \times M$ matrix:

$$T_1[i, j] = \begin{cases} 1, & when\ i = (n-1)H + k\ and\ j = n\ for\ n = 1, \ldots, M^2 \\ 0, & otherwise. \end{cases}$$

For instance, $n$ = 840 ($M$ = 50) is initially represented by vector $a_0$ whose 840th element is the only one while the rest are zero. $u$ (= $T_1 a_0$) is subsequently fed into $w^T$, resulting in $z$ (= $w^t u$) in which $z[40]$ = 1 and $z[50 + 21]$ = 1—denoting 40 and 21, respectively. These two vectors are merged through operator $T_2$ into $a_1$ ($\in \mathbb{Z}^M$; $a_1 = z[1:M] + z[M+1:2M]$). $T_2$ is, therefore, an $M \times 2M$ matrix:

$$T_2[i, j] = \begin{cases} 1, when\ j = i\ for\ i = 1, \ldots, M \\ 1, when\ j = i + M\ for\ i = 1, \ldots, M. \\ 0,\ otherwise. \end{cases}$$

This operation confers 1 on $a_1[21]$ and $a_1[40]$ in $a_1$. The address of each element represents a factor, and the element values its exponent so that the result of the first factorization is $21^1 \times 40^1$. Insofar as $a_1$ differs from $a_0$, the same cycle is repeated. Note that $a_1[1]$ (exponent of 1) is set to zero because a factor of 1 is redundant in factorization. The following cycle factorizes 21 and 40 in parallel, providing $a_2$ in which $a_2[2] = 1$, $a_2[3] = 1$, $a_2[7] = 1$, and $a_2[20] = 1$, i.e. $2^1 \times 3^1 \times 7^1 \times 20^1$.

**Direct search factorization**. Integer $n$ is repeatedly divided by a series of divisors (decreasing by one) until zero remainders. The first divisor is $\lfloor \sqrt{n} \rfloor$ that indicates the floor function (the largest integer small than or equal to $\sqrt{n}$). If the remainder is nonzero, $\lfloor \sqrt{n} \rfloor - 1$ is taken as the next divisor. With zero remainder, two factors (divisor and quotient) are obtained, and each factor is separately subject to the same factorization as above.

## Acknowledgements

D.S.J. and C.S.H. acknowledge the Korea Institute of Science and Technology Open Research Program (grant no. 2E27331).

## Author contributions

D.S.J. conceived the algorithm and its applications to arithmetic, supervised the entire work, and wrote the manuscript. C.S.H. modified the algorithm and refined the manuscript. G.K. conceived the algorithm, wrote the code, and analyzed the data. V.K. refined the algorithm, and D.K., I.K., J.K., H.C.W., and J.H.K. supported the technical detail of algorithm execution. All authors discussed the results and contributed to the refinement of the manuscript.

## Additional information

**Competing financial interests:** The authors declare no competing financial interests.

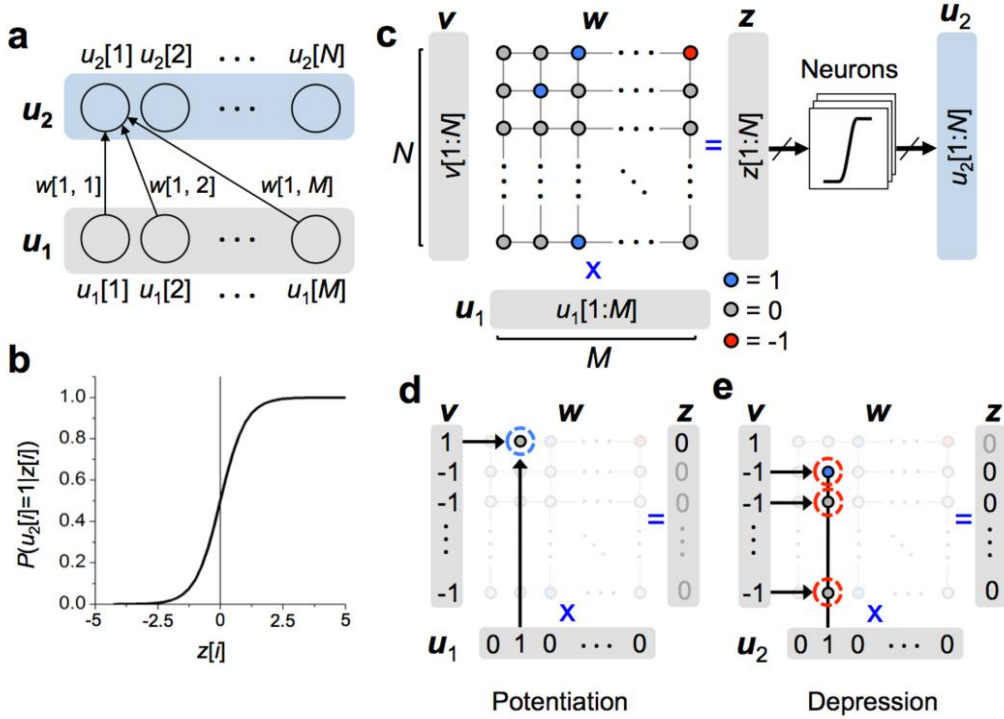

**Figure 1. MCHL algorithm working principle**. (**a**) Basic network of *M* input and *N* output binary stochastic neurons ($u_1$ and $u_2$: their activity vectors). (**b**) Behavior of $P(u_2[i] = 1)$ with z[i] when $a[i] = 0$. (**c**) Graphical description of the weight matrix *w* ($w \in \mathbb{Z}^{N \times M}$; $w[i,j] \in \{-1, 0, 1\}$) that determines the correlation between the input activity $u_1$ ($u_1 \in \mathbb{R}^M$; $0 \leq u_1[i] \leq 1$) and output activity $u_2$ ($u_2 \in \mathbb{Z}^N$; $u_2[i] \in \{0, 1\}$). This weight matrix *w* evolves in accordance to given pairs of an input $u_1$ and write vector *v* ($v \in \mathbb{Z}^N$; $v[i] \in \{-1, 1\}$), ascertaining the statistical correlation between $u_1$ and *v* by following the sub-updates. (**d**) **Potentiation**: a weight component at the current step *t* ($w_t[i, j]$) has a nonzero probability to gain +1 (i.e. $\Delta w_t[i, j] = 1$) only if $u_1[j] \neq 0$, $v[i] = 1$, and $w_t[i, j] \neq 1$; for instance, given $u_1 = (0, 1, 0, …, 0)$ and $v = (1, -1, -1, …, -1)$, $w_t[1, 2]$ has a probability of positive update. (**e**) **Depression**: all components $w_t[i, 2]$ ($i \neq 1$) are probabilistically subject to negative update (gain -1) insofar as $u_1[2] \neq 1$, $v[i] = -1$, and $w_t[i, 2] \neq -1$.

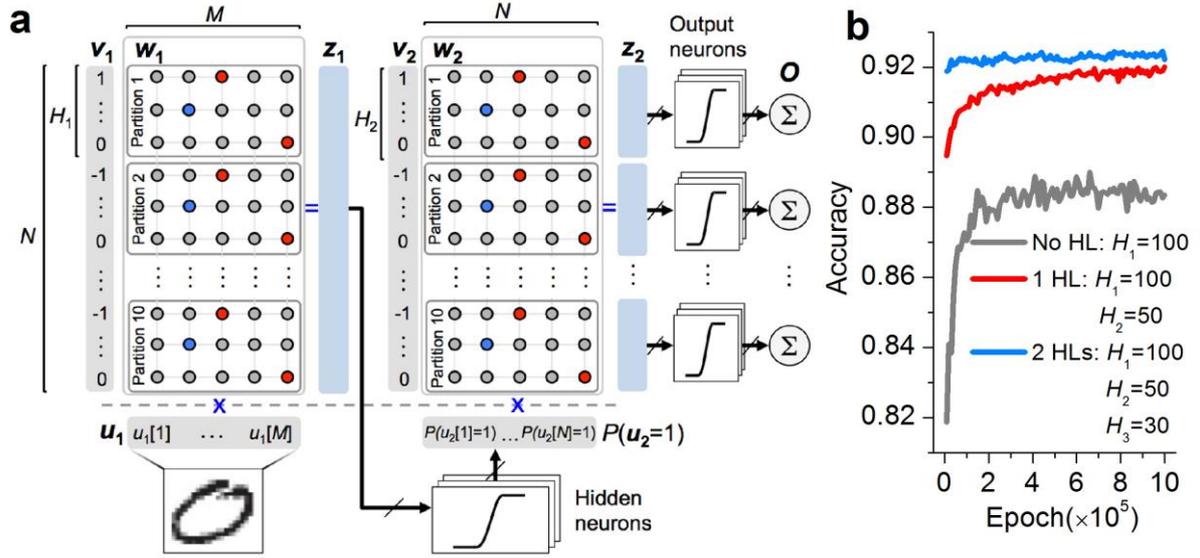

**Figure 2. Application to handwritten digit recognition**. (**a**) Schematic of the network architecture for handwritten digit recognition. A single HL is included. The matrix $w_1$ first maps the input vector $u_1$ to the hidden neurons. The probability that $u_2[i] = 1$ for all $i$'s is taken as an input vector to $w_2$ that maps the input vector to the output neurons. The write vector $v_1$ has 10 (the number of labels) buckets, each of which has $H_1$ elements, i.e. $N = 10H_1$. Each thick arrow indicates an input vector to a group of neurons (each neuron takes each element in the input vector) (**b**) Classification accuracy change in due course of training with network depth ($H_1 = 100$, $H_2 = 50$, $H_3 = 30$). $P_+^0$, $P_-^0$, and $\tau$ were set to 0.1, 0.1, and 1, respectively.

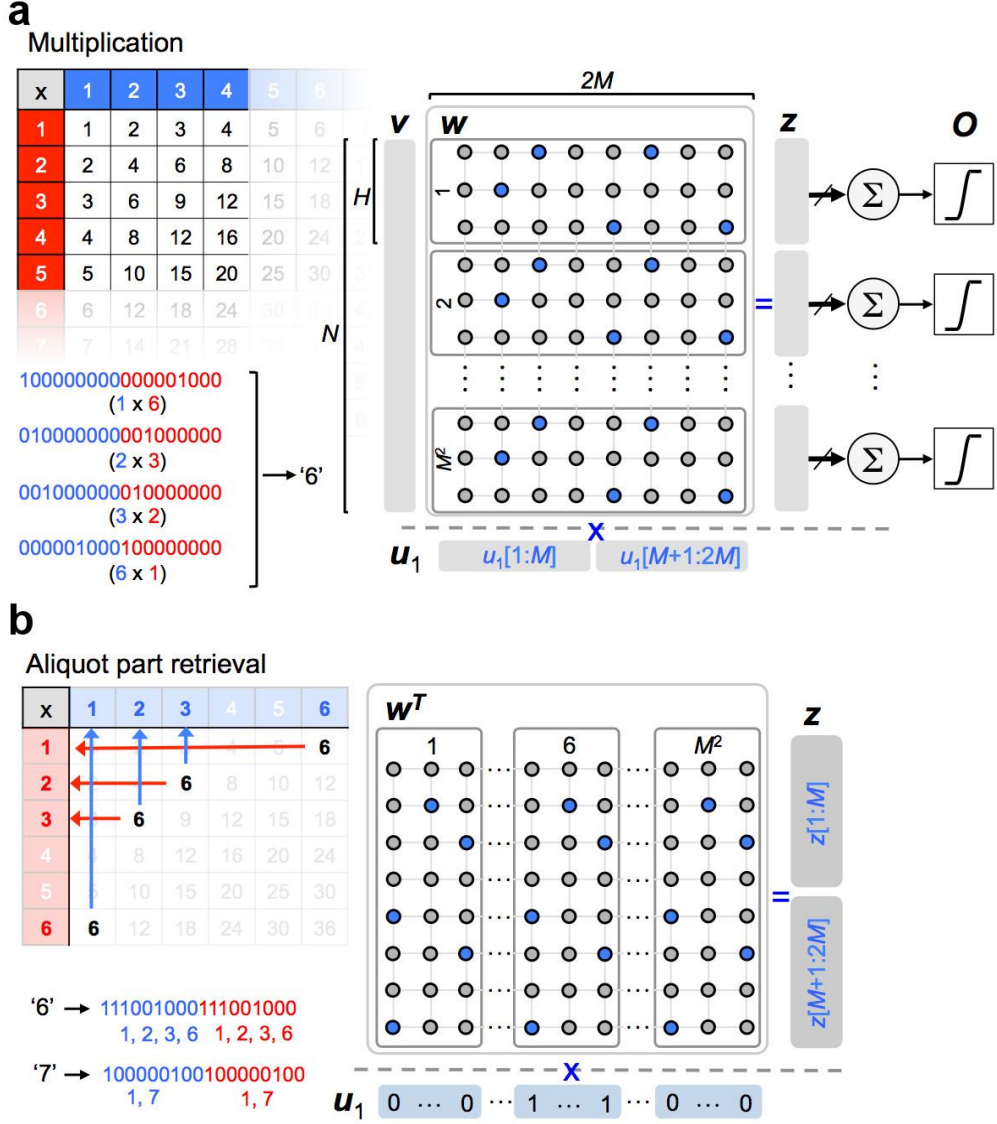

**Figure 3. Multiplication table memorization and aliquot part retrieval**. (**a**) Network architecture for multiplication table memorization. The numbers in the range $1 - M$ are described by one-hot vectors. Any two of total $M^2$ numbers are combined to form an input vector $u_1$ ($u_1 \in \mathbb{Z}^{2M}$; $u_1[i] \in \{0, 1\}$); for instance, when $M = 9$, $u_1$ for one and six is [100000000|000001000], where the first and last 9 bits indicate one and six, respectively, as shown in the figure. The correct answer serves as the label of chosen numbers; there are $M^2$ labels in total. Each label (bucket) has $H$ elements so that the write vector $v$ is a $M^2H$ long vector that is adjusted given the correct label. Given entire pairs of numbers in the table and their multiplication results, the matrix $w$ ($w \in \mathbb{Z}^{M^2H \times 2M}$) is adjusted. $P_+^0$, $P_-^0$, $a[i]$, and $\tau$ were set to 1, 0, 3, and 0.001, respectively (**b**) Network architecture of aliquot part retrieval

given the matrix $w$. The transpose of $w$ ($w^T$) finds the entire aliquot parts of a given number in a parallel manner in place. For instance, for number '6', an input vector $u_1$ ($M^2H$ long vector) has a single nonzero bucket (6th bucket) that is filled with ones. The output vector $z$ is [111001000|111001000], indicating the sum of four one-hot vectors ('1' + '2' + '3' + '6')—each of them is an aliquot part of 6. For prime numbers, the output vector includes only two 1's (1 and its own number) so that prime numbers can readily be found; for instance, 7 results in [100000100|100000100] as shown in the figure.

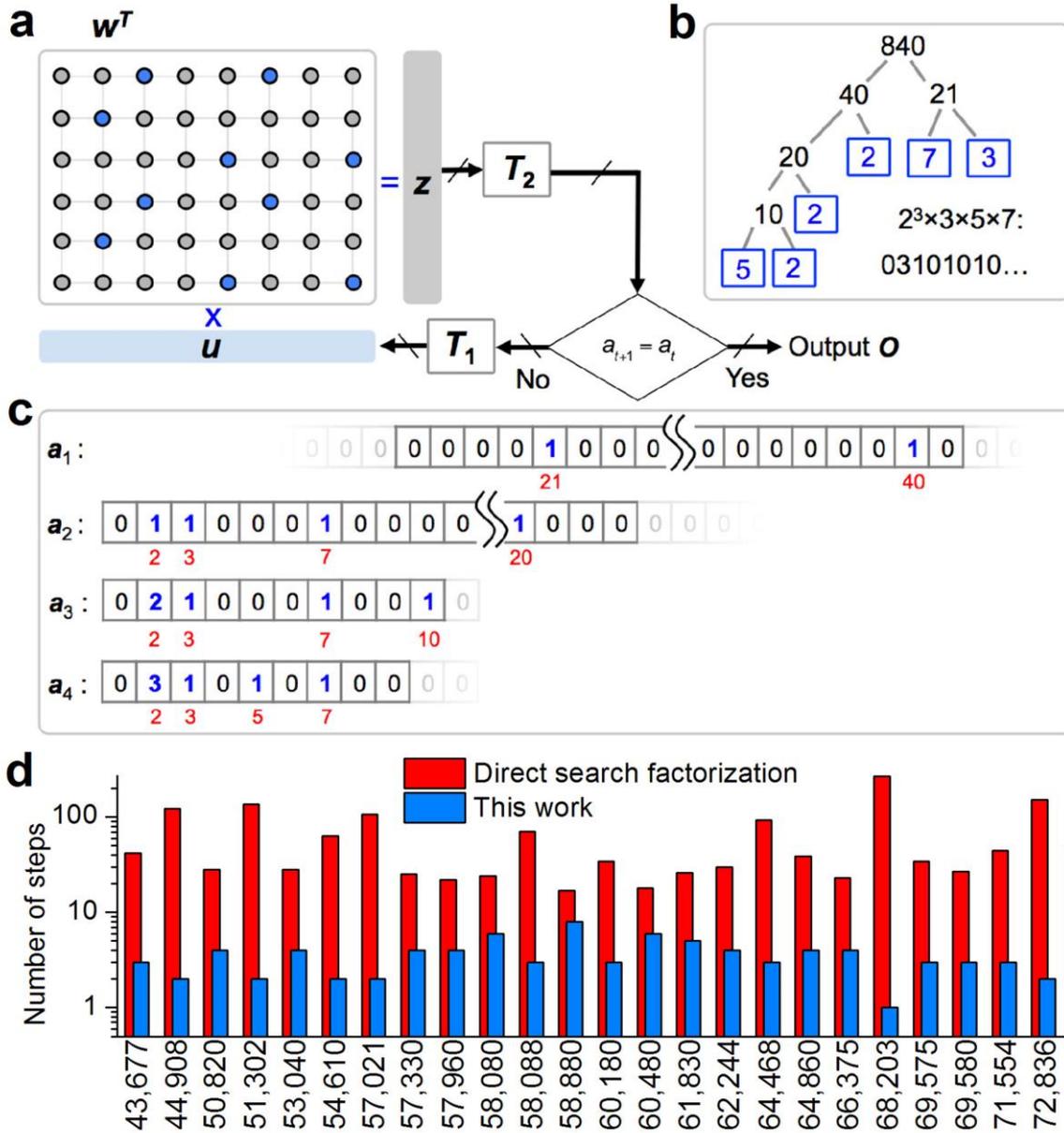

**Figure 4. Prime factorization.** (**a**) Memory ($w^T \in \mathbb{Z}^{2M \times M^2H}$) based iterative and parallel search for prime factors. Given an input vector $u$ standing for a certain number $n$, the matrix multiplication $w^T u$ outputs vector $z$ ($z \in \mathbb{Z}^{2M}$; $z[i] \in \{0, 1\}$) that reveals one pair of its factors—except 1 and itself—$z[1{:}M]$ and $z[M{+}1{:}2M]$ whose product yields $n$. Operator $T_2$ adds these two one-hot vectors, resulting $a_t$ ($a_t \in \mathbb{Z}^M$). The iteration terminates upon no further change in $a$ other than $a[1]$. Otherwise, operator $T_1$ transforms $a_t$ to $u$, and the next cycle continues. (**b**) Prime factorization of 840 = $2^3 \times 3 \times 5 \times 7$ with a matrix $w^T$ ($M = 100$, $H = 30$). The first iterative step outputs $a_1$ in (**c**); the address of each element indicates a factor, e.g. the 21st element, $a[21]$, means a factor of 21, and the element

value its exponent. Only $a_1[21]$ and $a_1[40]$ in $\boldsymbol{a}_1$ except $a_1[1]$ are nonzero, indicating 21×40. The second iteration outputs $\boldsymbol{a}_2$ whose nonzero elements are $a_2[2]$, $a_2[3]$, $a_2[7]$, and $a_2[20]$ (= 1, 1, 1, and 1, respectively), implying $2^2$×10×21. The third iteration respectively sets $a_3[2]$, $a_3[3]$, $a_3[7]$, and $a_3[10]$ to 2, 1, 1, and 1, i.e. $2^2$×3×7×10. The forth iteration sets $a_3[2]$, $a_3[3]$, $a_3[5]$, and $a_3[7]$ to 3, 1, 1, and 1, i.e. $2^3$×3×5×7 and an additional iteration does not alter other elements than $a[1]$ such that the prime factorization is completed. (**d**) The number of factorization steps until prime factors for 25 different integers in a 300 × 300 multiplication table. The results are compared with the direct search factorization.

**Table 1. Requirements for the update of nonzero probability**

| $v[j]$ | 1 | -1 |
|---|---|---|
| $u_1[j]$ | $0 < u_1[j] \leq 1$ | $0 < u_1[j] \leq 1$ |
| $w_t[i,j]$ | $\neq 1$ | $\neq -1$ |
| $u_2[i]$ | $\neq 1$ | $\neq 0$ |
| $P$ | $u_1[j]P_+^0$ | $u_1[j]P_-^0$ |